\documentclass[aps,prl,reprint,superscriptaddress,amsmath,amssymb,floatfix]{revtex4-2}
\usepackage{graphicx}
\usepackage{dcolumn}
\usepackage{bm}
\usepackage{booktabs}
\usepackage{multirow}

\begin{document}

\title{Hybridizing Equilibrium Propagation with Ising Machines for Efficient Energy-Based Learning}

\author{Chen-Rui Fan}
\affiliation{School of Artificial Intelligence, Beijing Normal University, Beijing 100875, China}
\author{Bo Lu}
\email{lubo@mail.bnu.edu.cn}
\affiliation{Laboratory for Advanced Computing and Intelligence Engineering, Information Engineering University, Zhengzhou 450001, China}
\author{Xing-Yu Wu}
\affiliation{School of Artificial Intelligence, Beijing Normal University, Beijing 100875, China}
\author{Tie-Jun Wang}
\affiliation{School of Physical Science and Technology, Beijing University of Posts and Telecommunications, Beijing 100876, China}
\author{Chuan Wang}
\email{wangchuan@bnu.edu.cn}
\affiliation{School of Artificial Intelligence, Beijing Normal University, Beijing 100875, China}

\date{\today}

\begin{abstract}
The rapid evolution of artificial intelligence has led to substantial advances in deep neural networks. Nonetheless, conventional GPU-based training remains highly energy-demanding, motivating the exploration of physical dynamics and compatible energy-based learning schemes, such as equilibrium propagation (EP). EP-based training, however, frequently suffers from convergence to local minima due to phase-space contraction. Here we introduce an Ising-dynamics-inspired equilibrium-propagation framework in which dissipative Hopfield relaxation is replaced by an extended phase-space dynamics with conjugate variables. The resulting training paradigm keeps the local two-phase learning rule of EP while changing the physical route by which neural states reach equilibrium. We show that this dynamics lowers effective energy barriers, accelerates convergence, improves noise robustness, and trains deep convolutional Hopfield networks on MNIST, FashionMNIST, and CIFAR-10 with performance comparable to backpropagation.
\end{abstract}

\maketitle

\paragraph{Introduction}--- The rapid development of artificial intelligence (AI), including large language models, autonomous driving, and smart healthcare, has driven the widespread use of deep neural networks such as CNNs \cite{lecun1998gradient}, ResNet \cite{he2016deep}, and Transformer \cite{vaswani2017attention}. These models are typically trained on large-scale GPU clusters, a process that is increasingly time-consuming, energy-intensive, and constrained by fundamental limitations \cite{xu2018scaling, patterson2021carbon,lillicrap2020backpropagation}. To overcome these bottlenecks, various physical hardware solvers have been explored for accelerating neural network applications \cite{xia2019memristive,aguirre2024hardware,jain2025heterogeneous,lin2018all,xu202111,shastri2021photonics,pai2023experimentally,wright2022deep,grollier2020neuromorphic}. Although these non-traditional platforms can achieve performance comparable to electronic computing, they are often pre-trained by backpropagation (BP) on external electronics before deployment, making them vulnerable to hardware noise and fabrication imperfections. This motivates renewed interest in physically grounded gradient-descent methods, where physical dynamics propagate gradients and the energy function governs learning \cite{hopfield1984neurons}. Such methods enable inference and training through local learning rules, offering advantages for physical systems \cite{stern2023learning}. A representative example is Equilibrium Propagation (EP), an energy-based learning algorithm for Hopfield networks that unifies training and inference within the same dynamical process \cite{scellier2017equilibrium}. By applying target-controlled external energy to the output layer, EP implicitly propagates learning signals through hidden layers, approximating BP gradients while alleviating limitations of conventional physical training methods \cite{scellier2019equivalence, ernoult2019updates,10.5555/3600270.3601211,helwegen2019latent,laborieux2021scaling, laydevant2021training}, and has been extended to diverse hardware platforms \cite{yi2023activity,wang2024training, rageau2025training,laydevant2024training}.

However, the dynamics in Hopfield networks are non-Hamiltonian dissipative systems\cite{scellier2017equilibrium}, governed by $\dot{x}=-\nabla E(x)$. The evolution vector field in such systems exhibits negative divergence, leading to the phase space contraction and a tendency to collapse into an attractor. This characteristic often results in the network falling into local energy minima, which undermines the overall effectiveness of training.
Recent progress has shown that the structural correspondence between Hopfield networks and the Ising model enables neural network training to be interpreted as Hamiltonian minimization, providing a physical foundation for Ising-machine-inspired learning\cite{yamamoto2017coherent,mohseni2022ising}. However, directly mapping multilayer network parameters onto large-scale spin systems faces severe scalability issues, including prolonged convergence and susceptibility to local optima\cite{laydevant2024training,gower2025learning,srivastava2015training}.
Inspired by the ability of a simulation bifurcation machine (SB) based on a nonlinear oscillator network Hamiltonian system to simulate the quantum adiabatic evolution process\cite{goto2019combinatorial,kanao2021high,goto2021high,razmkhah2024josephson,yamaji2023correlated,ucpinar2024scalable}, we propose a hybrid neural network training framework inspired by Hamiltonian dynamics and equilibrium propagation. By extending simulated bifurcation to a continuous simulated bifurcation machine (cSB), we treat energy-based neural networks as continuous Ising-like dynamical systems and exploit their low-energy state search to accelerate equilibrium convergence. Experiments show that cSB-EP achieves lower energy distributions, faster convergence, improved training accuracy, and stronger robustness to noise than conventional EP. Furthermore, cSB-EP scales to CNNs, attaining accuracy comparable to backpropagation on CIFAR-10 while achieving several-fold speedup over existing EP algorithms. These results suggest that cSB-EP provides not merely an application of Ising machines to neural networks, but a new Ising-machine-dynamics-inspired training paradigm for efficient, scalable, and hardware-compatible energy-based learning.

\paragraph{Ising Machine dynamics-inspired equilibrium propagation} The Hopfield network, based on the energy function $E$, could be expressed as:
\begin{equation}
	E_{h} = \frac{1}{2}\sum_{i} s_{i}^{2} - \sum_{i>j} W_{ij} \rho(s_{i}) \rho(s_{j}) - \sum_{i} h_{i} \rho(s_{i})\label{eqEP_energy},
\end{equation}
$s$ represents the states of the neurons, $W_{ij}$ is the synaptic weights, $h_{i}$ denotes the neuron biases, and $\rho$ is the nonlinear activation function. In EP, by fixing the pattern $v$ to be learned, the mean squared error $C =\beta\|v-s\|^2$ can typically be used as the loss function. The energy function $E_{h}$ is minimized to achieve the memorization of the pattern.
For a network with parameters $\theta$, we denote the neuron state as \(s_{\theta,v}^\beta\), which corresponds to the fixed point under the pattern \(v\) and the perturbation parameter \(\beta\). The minimum energy is denoted as \( E_{h}(\theta,v,s_{\theta,v}^\beta)\). Consequently, the neuron dynamics follows the condition ${\partial E_{h}}/{\partial s} = 0$ and $\arg \min C(v, s)$, which can be expressed as:
\begin{equation}
	\dot{s}_i = - s_i + \sum_{j=1}^N W_{ij} \rho(s_j)\rho'(s_j).
\end{equation}
The above equation indicates that the system has reached a stationary point of the energy function. This process is similar to the ground state search in the Ising model and can be implemented in various approaches. 
However, this problem is NP-hard because the number of possible spin configurations grows exponentially with system size. Consequently, no efficient algorithms exist for large-scale instances.

An optimization algorithm that leverages the adiabatic evolution of classical nonlinear Hamiltonian systems employs a network of Kerr nonlinear parametric oscillators (KPOs) to efficiently simulate the Ising problem quantum-adiabatically. This approach, termed the Simulated Bifurcation (SB) machine, is described by the system Hamiltonian:
\cite{goto2019combinatorial}:
\begin{align}
	H_q(t)
	&= \hbar \sum_{i=1}^{N}
	\Biggl[
	\frac{K}{2}\, a_i^{\dagger 2} a_i^{2}
	-\frac{p(t)}{2}\,\bigl(a_i^{\dagger 2}+a_i^{2}\bigr)
	+\Delta_i\, a_i^{\dagger} a_i
	\Biggr] \nonumber\\
	&\quad - \hbar \sum_{i=1}^{N}\sum_{j=1}^{N} J_{ij}\, a_i^{\dagger} a_j .
\end{align}
where $a_i^{\dagger}$ and $a_i$ are the creation and annihilation operators, respectively. $p(t)$ is the time-dependent pumping amplitude, $K$ represents the nonlinear interaction strengths, $\Delta_i$ is the detuning frequency between the resonance frequency of the $i$th KPO and the pump, and $J_{ij}$ is the coupling strength of the $i$th and $j$th spins. 
Through quantum adiabatic bifurcation, each KPO evolves into a binary state. The corresponding Ising spin in the ground state corresponds to the sign of the KPO's final amplitude. In this process, the Hamiltonian of the target Ising model is encoded within the coupling terms, which directly govern the outcome of the evolution.
However, the expectation value of $a_i$ is approximated as a complex amplitude $x_i + iy_i$ by classical approximation. 
The classical mechanical Hamiltonian and the equations of motion for this system can be simplified as follows
\begin{align}
	\dot{y}_i &= -[1 - p(t)]x_i + \sum_{j=1}^N J_{i,j}x_j \label{eq:y_dynamics}  \\
	\dot{x}_i &= y_i, \quad \left| x_i \right| \le 1. \label{eq:x_dynamics}
\end{align}
The above approach is designed for a binary spin system $s = sgn(x) \in \{\pm 1\}$, and does not apply to current neural networks. This involves several challenges, notably the unsuitability of activation functions, the difficulty in optimizing discrete values, and a constrained capacity for information loading all of which present significant difficulties.

Therefore, based on the Hamiltonian system above, we propose a continuous Simulated Bifurcation machine (cSB) for neural network optimization. To derive its dynamics, we consider an oscillator under phase noise compression and neglect higher-order terms. The activation function $\rho(x) = \min \!\bigl(\max(x,0),1\bigr)$ is used to replace the inelastic potential barrier, and accelerates the computation. Simultaneously, the gain term is omitted, and a damping term is introduced to ensure rapid convergence. This yields the following dynamical equations for the cSB:
\begin{align}
	\dot{sy}_i &= - s_i - \gamma sy_i + \rho(\sum_{j=1}^N J_{ij} s_j), \\
	\dot{s}_i &= sy_i.
\end{align}

Meanwhile, the conjugate coordinate $sy$ introduced by the Hamiltonian system provides substantial inertial momentum during the initial phase of evolution. This momentum enables the $s$ neurons to overcome shallow local minima in the energy landscape, regardless of the training stage. To understand this, consider the time evolution of the potential functions for two types of systems. For a non-Hamiltonian dissipative system, let $V_{1}(x)=0.5\,x^{\top}x-x^{\top}Jx$, which satisfies $\mathrm{d}V_{1}(x)/\mathrm{d}t=(\nabla V_{1}(x))^{\top}\dot{x}=-\lVert\nabla V_{1}(x)\rVert^{2}<0$. Consequently, the system energy decreases strictly monotonically, forcing the dynamics to converge only to gradient-zero points and making it prone to being trapped in local minima.

In contrast, for the Hamiltonian system, consider the extended potential function $V_{2}(x)=0.5\,y^{\top}y+0.5\,x^{\top}x-x^{\top}Jx$. Its gradient along the trajectory satisfies $\mathrm{d}V_{2}/\mathrm{d}t=\partial V_{2}/\partial x\cdot\dot{x}+\partial V_{2}/\partial y\cdot\dot{y}=(x-Jx)^{\top}y+y^{\top}(x-Jx)-\gamma y^{\top}y$. At the stable points of the non-Hamiltonian dissipative system, this derivative remains negative. However, the inertial term $sy$ allows the system to temporarily move into regions of higher potential energy, thereby facilitating escape from local minima.

To investigate the effect of introducing the coupling term $sy$ on the system escape out of local minima, we set $a_\delta=(a_x,y_a(\delta))$ and $s_\delta=(s_x,y_s(\delta))$ be the stationary points of $V_{\delta}$ corresponding to $a_x$ and $s_x$, respectively. When $\delta \to 0^+$, according to the Freidlin-Wentzell principle \cite{freidlin1998random}, we have $\Delta V_\delta<\Delta V_0:=V_1(s_x)-V_1(a_x)$. Here, $\Delta V_0$ and $\Delta V_\delta$ represent the quasi-potential differences between $s_\delta$ and $a_\delta$ for the uncoupled and weakly coupled cases, respectively. 
Therefore, the probability of a weakly coupled system escaping from a local optimum satisfies \cite{bovier2016metastability,bovier2004metastability} $k_\delta /k_0 \propto exp(\frac{\Delta V_0-\Delta V_\delta}{\varepsilon})$ and validated on a three-spin Ising model(Refer to the END MATTER).

The central change is therefore not only a hardware substitution, but a change of training dynamics: the free phase and the nudge phase of EP are both solved through an extended phase-space evolution rather than through purely dissipative Hopfield relaxation. The local learning rule is retained, while the route to equilibrium is altered by the conjugate coordinate.

\begin{figure}[t]
	\centering
	\includegraphics[width=\columnwidth]{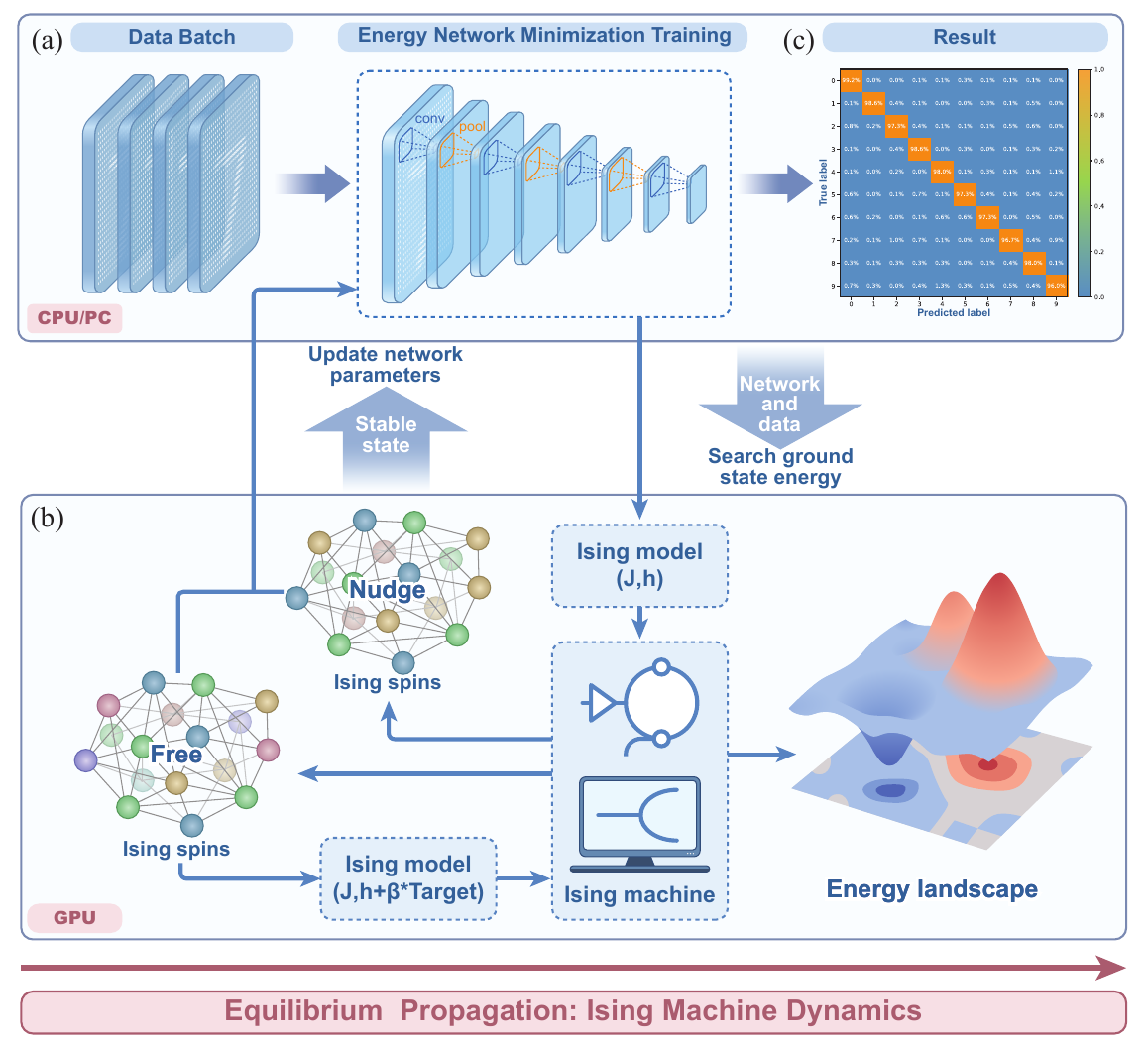}
	\caption{\textbf{Ising machine dynamics-inspired training paradigm for energy-based neural networks.} (a) The algorithm is implemented by combining CPU and GPU. The dataset is integrated in batches and loaded into a CNN on the CPU. Network parameters are updated in the CPU by the stable state of the two phases. (b) The network evolution toward the ground state is simulated using the cSB algorithm, with neuron states computed via a GPU-based implementation. The recognition result is determined by the stable state of the free phase. (c) The network recognition results are displayed in a confusion matrix, with the vertical axis representing the predicted outcomes of the neural network and the horizontal axis representing true targets.}
	\label{fig1}
	\label{fig:paradigm}
\end{figure}

As illustrated in Fig. \ref{fig1}, our design integrates EP with cSB. Input data processing and network parameter updates are performed on the CPU/PC, while the two fixed-point convergence stages are accelerated by the GPU with the aid of an Ising machine. This significantly reduces the overall computation time. Given the input data and initial network parameters, cSB calculates and stores the ground-state neuron configurations, completing the free phase as shown in Fig. \ref{fig1}(b).

Subsequently, a perturbation is applied to the network output layer to drive the system toward the target state, enabling supervised training. The perturbation strength is controlled by the parameter $\beta$. The cSB computation then yields the neuron states for the nudge phase. The network weights are updated on the CPU/PC according to the EP learning rule, which requires multiple iterations. Finally, by comparing the test set outputs with their targets, we obtain the confusion matrix shown in Fig. \ref{fig1}(c). Here, deeper red shades indicate higher accuracy. The strong diagonal results demonstrate that our network architecture successfully achieves effective training on the MNIST dataset.

\paragraph{Energy-landscape mechanism and training behavior} This energy-based neural network is centered around an energy function and utilizes learning rules based on a set of stored patterns. Through dynamic neuronal updates, the network monotonically converges to local energy minima, thereby achieving supervised learning. The cSB-EP-based architecture enables a faster search for the ground-state energy. Accordingly, we investigated the energy distribution of networks trained under cSB-EP with supervised learning. Specifically, we train a fully connected neural network with one hidden layer on the MNIST dataset using the cSB-EP framework.

\begin{figure}[t]
	\centering
	\includegraphics[width=\columnwidth]{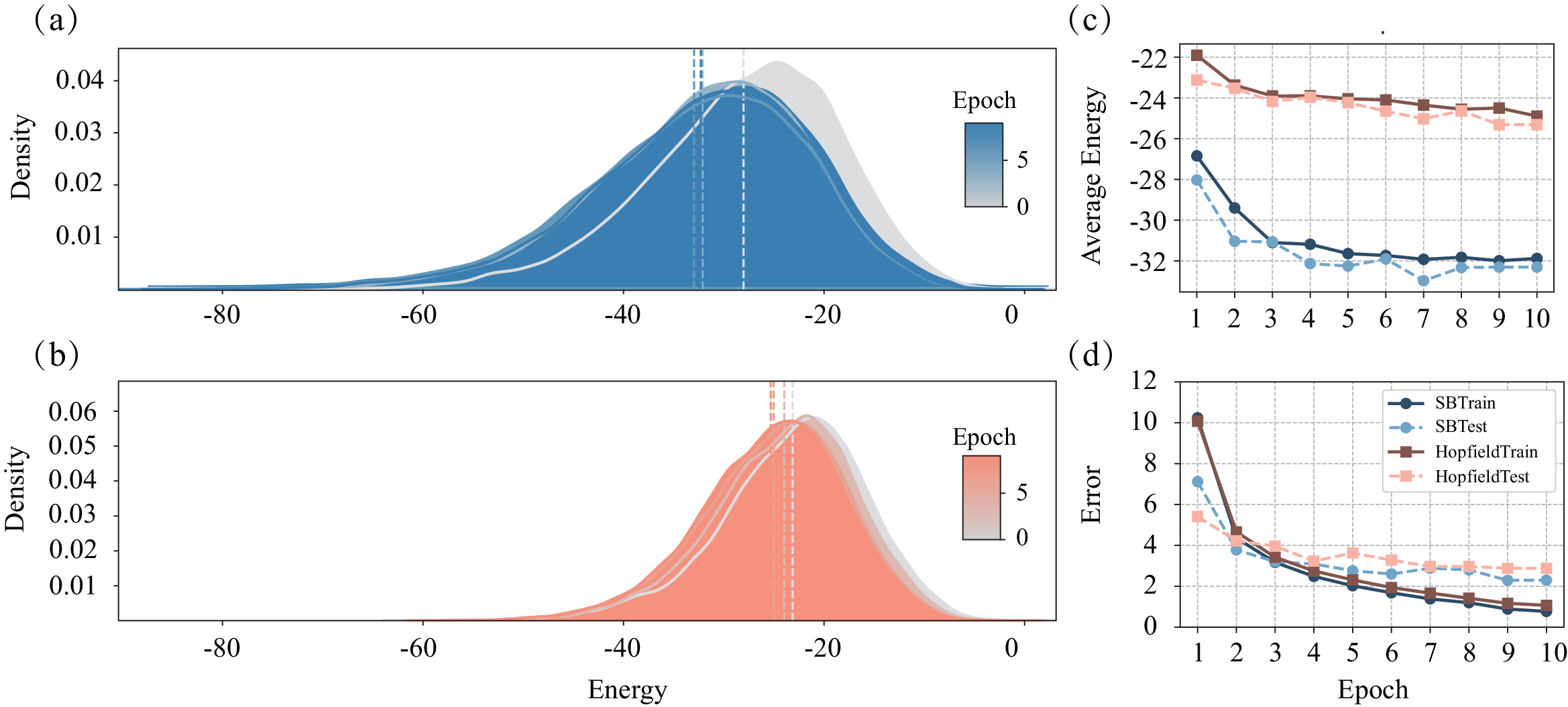}
	\caption{\textbf{Energy distribution during network training.} (a) and (b) show the energy changes of the network by training with cSB-EP and EP, respectively. The energy distribution maps with varying shades represent different epochs, and the vertical dashed lines indicate the energy mean. (c) shows the average energy of the network during training across epochs. (d) shows the error in recognition results across epochs. The red line represents the EP, and the blue line represents the cSB-EP. Solid lines represent the training set, and dashed lines represent the test set.}
	\label{fig3}
	\label{fig:energy}
\end{figure}

To investigate the effects of different dynamical systems on network training, we analyzed the network energy throughout the training process, as shown in Fig. \ref{fig3}. The trained model was a fully connected neural network with a single hidden layer of 120 neurons. Training was performed on 60,000 handwritten digit images of size 28$\times$28, with a batch size of 128. Each batch underwent 20 free-phase and 15 nudge-phase iterations. Over 10 training epochs, we recorded the energy distribution for each image as it evolved to a stable state under both cSB-EP and conventional EP, shown in Fig. \ref{fig3}(a) and (b), respectively. Compared to conventional EP, cSB-EP achieves a lower energy distribution and accelerates the shift of training energies toward lower values. Fig. \ref{fig3}(c) further presents the epoch-wise evolution of the average network energy on both training and test sets. The corresponding error rate variation is displayed in Fig. \ref{fig3}(d). Consistent with the energy-based analysis, cSB-EP exhibits a lower error rate and superior learning performance.

\begin{figure}[t]
	\centering
	\includegraphics[width=\columnwidth]{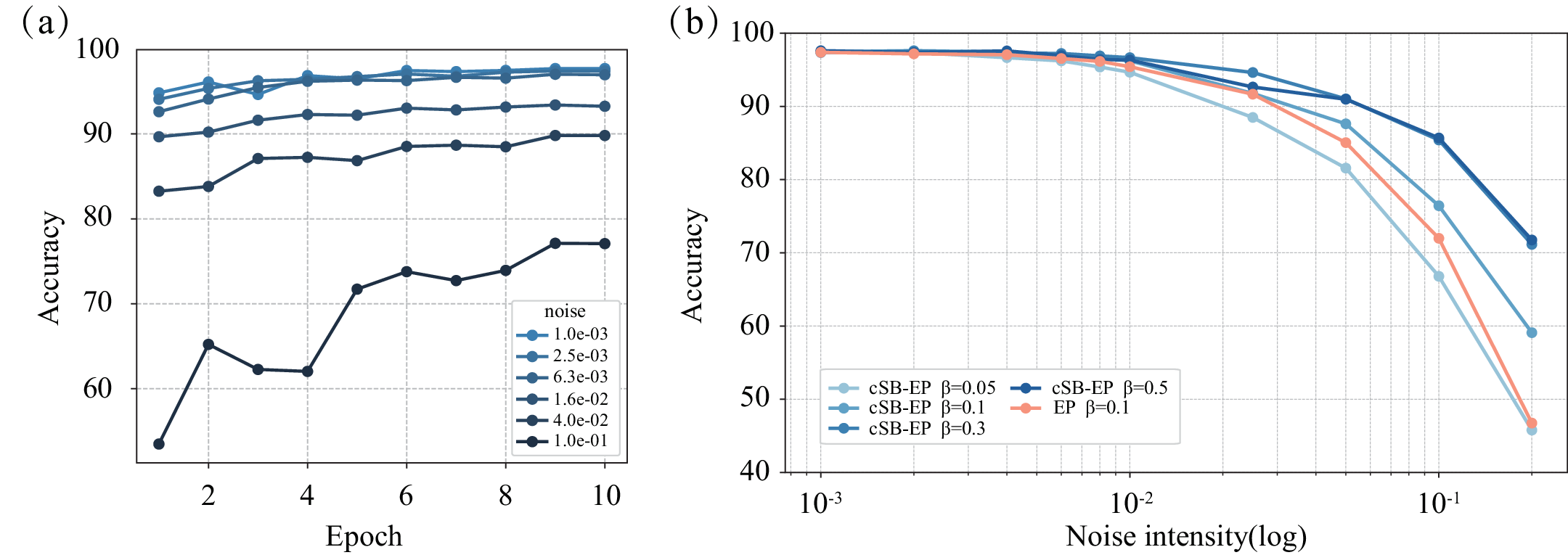}
	\caption{\textbf{Robustness of the Hopfield Network trained with cSB-EP under noisy conditions.} (a) The change in training accuracy across epochs for different intensity noise. (b) The accuracy of cSB-EP and EP across different intensity noise for various perturbation parameters $\beta$.}
	\label{fig4}
	\label{fig:noise}
\end{figure}

Due to factors such as limited floating-point bit width, analog-to-digital converters (ADCs), and the inherent noise properties of opto-electronic systems, noise is inevitably introduced during iterative computations in widely studied digital circuits or opto-electronic Ising machines, leading to computational errors. The robustness of the algorithm against noise determines its performance when extended to physical hardware.
To evaluate the noise tolerance of the neural network trained with cSB-EP, we injected zero-mean Gaussian noise of small-amplitude into the dynamics of each neuron in the network. The effect on the training performance is shown in the Fig.\ref{fig4}(a).
When the noise intensity is below \( 10^{-2} \), the classification accuracy remains above 90\%, with negligible degradation in network performance. However, when the noise intensity increases to \( 10^{-1} \), the ground state solution computed by SB deviates significantly from the true value. This results in substantial errors in the gradient direction learned by the network, causing the accuracy to drop to around 75\%. Notably, slight noise can be beneficial, as its randomness helps the search process escape local optima, whereas strong noise introduces large deviations that adversely affect training accuracy.

In EP, \( \beta \) is a crucial parameter that controls the amplitude of the pattern bias field in the nudge phase. Fig.\ref{fig4}(b) shows the training results under different \( \beta \) with noise. Under the same intensity noise, as \( \beta \) increases, the test accuracy also rises, indicating that the stronger the external perturbation, the stronger the recognition capability of the model. For the same \( \beta \), an increase in intensity noise interferes more with the evolution of neurons, making it difficult to compute the correct ground-state spin configuration, which leads to a decline in test accuracy. Overall, larger \( \beta \) exhibit stronger noise resistance, and for the same noise intensity, recognition ability decreases less. Under noisy conditions, increasing the strength of external perturbation can enhance the network learning ability towards the pattern direction, reducing errors during neuron training and partially mitigating the impact of noise on network learning. However, this effect does not increase indefinitely; once a certain threshold is reached, the test accuracy no longer increases with \( \beta \). To achieve optimal performance in EP, a small \( \beta \) is preferred, but a low \( \beta \) requires low noise to maintain high accuracy. At the same perturbation with \( \beta = 0.1 \), cSB-EP shows a slight decrease in training accuracy compared to EP, demonstrating stronger robustness to noise.
The cSB-EP encodes knowledge through energy landscapes and performs retrieval via dynamics, serving as a bridge between classical memory models and deep learning.


\paragraph{Deep convolutional Hopfield networks} While EP has been widely validated on small-scale MLPs, more demanding tasks often necessitate advanced neural network architectures. In this section, we train a $5$-layer deep convolutional Hopfield network using cSB-EP and evaluate it on MNIST, Fashion MNIST, and CIFAR-10. The results are summarized in Table \ref{tab:EP_comparison}. Compared to prior work, we adopt a consistent convolutional kernel size and a network architecture comprising four convolutional layers and one fully connected classification layer, achieving marginally superior recognition performance. As shown in the comparative results, the proposed cSB-EP architecture attains the lowest test error on MNIST and performs closer to Backpropagation (BP) than conventional cEP on Fashion MNIST and CIFAR-10 datasets. These findings demonstrate that the cSB-EP architecture can be effectively extended to more complex Deep Convolutional Hopfield Networks.

\begin{table*}[!t]
	\centering
	\caption{Recognition Error Rates $(\%)$ of cSB-EP on MNIST, FashionMNIST and CIFAR-10 Classification Datasets.}\label{tab:EP_comparison}
	\begin{tabular}{llcccccc}
		\hline
		\multicolumn{2}{c}{} & \multicolumn{2}{c}{MNIST} & \multicolumn{2}{c}{FashionMNIST} & \multicolumn{2}{c}{CIFAR10} \\
		\hline
		\multicolumn{2}{c}{} & test & train & test & train  & test & train \\
		\hline
		This work (cSB-EP) & cEP (4-conv) & 0.4 & 0.06 & 6.32 & 3.41& 10.98 & 2.08 \\
		\hline
		\multirow{2}{*}{Laydevant et al. \cite{laydevant2021training}} & EP (2-conv) & 0.85 & 0.46 & - & - & - & -\\
		& EP (2-fc)& 2.28 & 0 & - & - & - & -\\
		\hline
		\multirow{2}{*}{Scellier et al.\cite{scellier2023energy}} & cEP (4-conv) & 0.44 & 0.20& 6.47 & 4.01 & 11.1 & 5.6 \\
		
		& BP (4-conv) & 0.42 & 0.23 & 6.12 & 3.09 & 10.1 & 3.1 \\
		\hline
		Labrieux et al.\cite{laborieux2021scaling} & cEP (4-conv) & - & -  & - & -& 11.68 & 4.98 \\
		\hline
	\end{tabular}
	\footnotetext{The recognition error rates of cSB-EP on database classification, with both training and testing error rates expressed as percentages. The results are compared with those in the existing literature on EP. cEP: Center-Symmetric EP}
\end{table*}

\begin{figure}[t]
	\centering
	\includegraphics[width=\columnwidth]{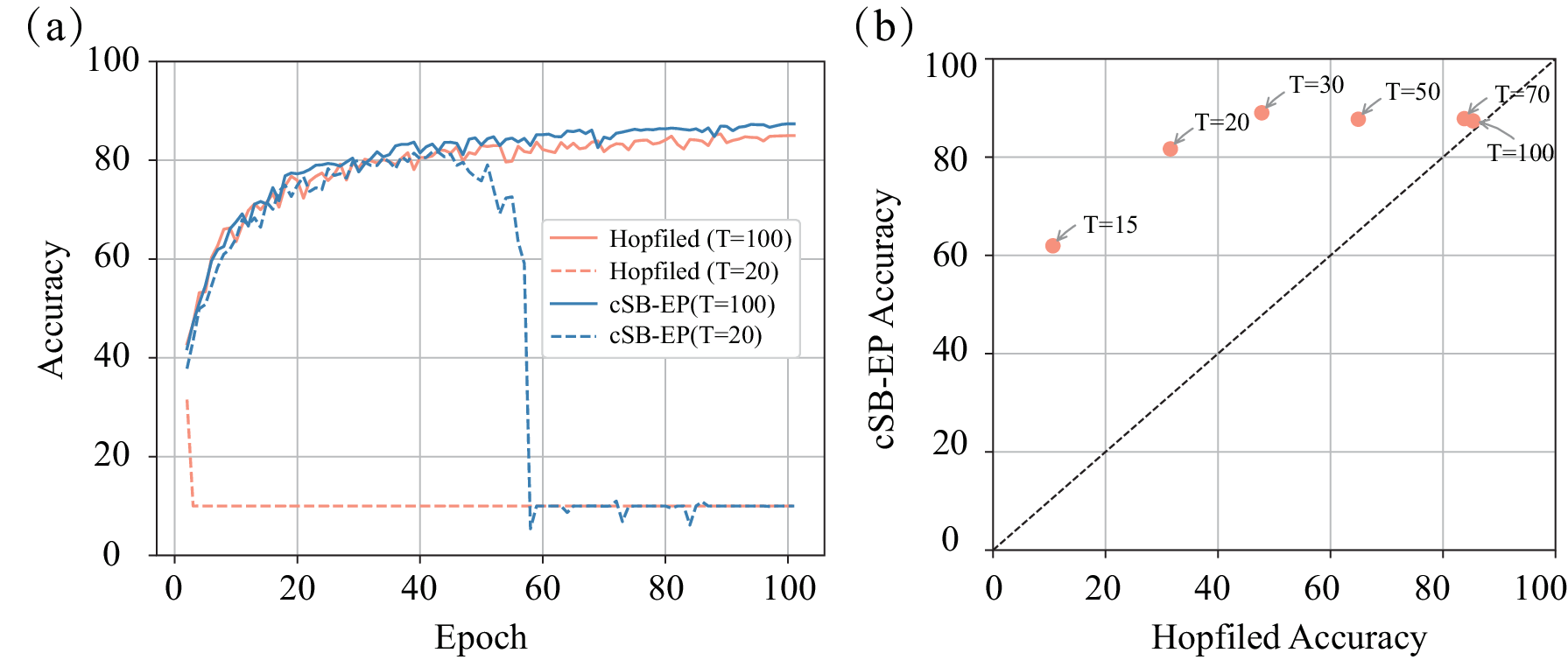}
	\caption{\textbf{Phase-stable iterations limit for energy neural networks.} (a) Accuracy variation with epochs under different iteration constraints (T = 20 and 100). (b) Comparison of accuracy between cSB-EP and EP at the end of training under the same iteration constraints. Iterations (T) are indicated in the figure.}
	\label{fig5}
	\label{fig:iteration}
\end{figure}

Based on the local minima transition rate analysis, cSB-EP achieves exponential acceleration in ground-state search compared to EP.  Therefore, we investigated the training accuracy of the deep convolutional Hopfield Network on the CIFAR-10 dataset under different iteration constraints for both cSB-EP and EP, as shown in Fig.\ref{fig5}. Initially, we examined the CNN training performance for iterations set to $20$ and $100$, as shown in Fig.\ref{fig5}(a). For the case of $T=100$ iterations, since the iterations allow the neural dynamics to converge to stability, both cSB-EP and EP were able to effectively complete the learning task. However, cSB-EP achieved slightly higher accuracy than EP.
In contrast, for $T=20$ iterations, where stable convergence could not be achieved, EP lost its learning capability early on due to the inability of the neurons to converge to the ground state under label perturbations, resulting in a drop in accuracy from $0.3$ to $0.1$. In comparison, cSB-EP, benefiting from the dynamical advantages introduced by weak coupling, quickly returned to the ground state under perturbations of simpler patterns, achieving an accuracy of up to 0.8. However, for more difficult patterns, the iteration failed to meet the stability convergence condition, and learning could not continue.

We compared the impact of different iterations on training accuracy, as shown in Fig.\ref{fig5}(b). Compared to EP, cSB-EP achieved effective learning with a minimal number of iterations of $T=15$, and stable convergence with saturated accuracy was reached around $T=30$. For the same accuracy, cSB-EP provided $3\times$ computational speedup over EP. Furthermore, for the selected iterations, all data points fall in the top-left part of Fig.\ref{fig5}(b), indicating that cSB-EP demonstrates superior learning performance under the same iteration conditions. Therefore, compared to existing EP algorithms, our results demonstrate a reduction in ground-state search iterations, leading to accelerated network training and computation, while achieving training performance comparable to BP. Additionally, the modular design of our solution enhances its integration with physical hardware, such as Ising machines, thereby accelerating the application of EP in AI.

\paragraph{Conclusion}---This study introduces a dynamic training scheme for energy-based neural networks using the cSB framework, which significantly enhances both convergence speed and computational accuracy. By constructing a Hamiltonian system via conjugate coordinates, we effectively mitigate the local minima problem inherent to non-Hamiltonian dissipative systems, such as classical Hopfield networks, caused by phase-space contraction. This enables an efficient search for minimum-energy configurations along non-self-intersecting trajectories. Compared to discrete-time EP and agnostic EP \cite{ernoult2019updates, scellier2022agnostic,scellier2023energy}, the cSB-EP achieves at least $3\times$ improvement in computational efficiency. Additionally, energy-based neural networks trained with cSB attain lower-energy spin configurations than those obtained through traditional EP, with benchmark tests further validating their superior training performance.

Furthermore, the cSB-EP scheme exhibits enhanced robustness to noise, a critical feature for Ising machines implemented on physical platforms such as analog circuits or opto-electronic systems. 
Combined with existing large-scale Ising machine systems — for example, optical systems\cite{gao2025all,honjo2021100}, opto-electronic systems\cite{bohm2019poor, cen2022large}, and electronic oscillators\cite{cilasun2025coupled}, we anticipate that this framework, we anticipate this framework to emerge as a leading Ising machine based neural training solution, offering a viable pathway toward energy-efficient, low-power AI deployment in edge computing and IoT devices.

\section*{End Matter}
\textit{Barrier reduction by weak coupling.} In a Hopfield Network system, the energy function is given by
\begin{equation}
	V_1(x)=\frac{1}{4}x^Tx^Txx-\frac{1}{2}x^Tx+x^TJx.
\end{equation}
Consider the potential function of a weakly coupled field
\begin{equation}
	V_{\delta}(x,y)=V_1(x)+V(y)+\delta C(x)y^2, 
\end{equation}
where $V(y)=\frac{1}{4}\beta y^4-\frac{1}{2}\alpha y^2$, $0<\delta \ll 1$, and $\alpha(\beta)>0$. This potential energy function adjusts $y$ to a more favorable position near the saddle point, thereby lowering the saddle point energy and consequently reducing the energy barrier.
According to the Freidlin-Wentzell principle \cite{freidlin1998random}, the quasi-potential (starting from the small well $A$) is defined as:
\begin{equation}
	V(A\to z)=\inf_{T>0}\inf_{\phi(0)\in A,\phi(T)=z}\frac{1}{4}\int_0^T\|\dot{\phi}+\nabla V(\phi)\|^2dt.
\end{equation}
For gradient systems, this simplifies to $V(a\to s)=V(s)-V(a)$, where $s$ is the minimum saddle point connecting the corresponding attraction domains.

Let $a_\delta=(a_x,y_a(\delta))$ and $s_\delta=(s_x,y_s(\delta))$ be the stationary points of $V_{\delta}$ corresponding to $a_x$ and $s_x$, respectively. So we have
\begin{eqnarray}
	&\partial_{y}V_{\delta}=V^{\prime}(y)+2\delta C(x)y,\\
	&[\beta y^{2}-\alpha+2\delta C(x)]y=0.
\end{eqnarray}
When $\delta\to0^+ $, $C(a_x)=0$, we have $y= 0$ or $\pm a$, and $a = \sqrt{\frac{\alpha}{\beta}}$.
It can be obtained using the implicit function theorem $y_a(\delta)=\pm a+O(\delta)$\cite{kato2013perturbation}.
Similarly, at the saddle point $x=s_x$, the relation $y_s(\delta)=\pm\sqrt{\frac{\alpha-2\delta C(s_x)}{\beta}}=\pm a\sqrt{1-\frac{\delta}{\alpha}C(s_x)}+O(\delta^{3/2})$ could be obtained.

Meanwhile, we have a quasi-potential that is defined as
\begin{equation}\begin{aligned}
		& V_{\delta}(a_{\delta})=V_1(a_x)+V(\pm a)+\delta C(a_x)a^2=V_1(a_x)+V_{\min} \\
		& V_{\delta}(s_{\delta})=V_1(s_x)+V\left(y_s(\delta)\right)+\delta C(s_x)y_s(\delta)^2.
\end{aligned}\end{equation}
By substituting $V(y)$, we have $V\left(y_s(\delta)\right)+\delta C(s_x)y_s(\delta)^2=V(\pm a)-\delta C(s_x)a^2+O(\delta^2)$.
Then, we have
\begin{equation}
	\Delta V_{\delta}=V_{\delta}(s_{\delta})-V_{\delta}(a_{\delta})=[V_{1}(s_{\delta})-V_{1}(a_{\delta})]-\delta C(s_{x})a^{2}+O(\delta^{2}).\end{equation}
Here, $V_{1}(s_{\delta})-V_{1}(a_{\delta})$ is the quasi-potential of $\Delta V_1(a_{\delta}\to s_{\delta})$. Thus, for any sufficiently small $\delta >0$, we have $\Delta V_{\delta}<\Delta V_1$.
For a given time window $T$, the Eyring-Kramers transition rate satisfies \cite{bovier2016metastability,bovier2004metastability}
\begin{equation}
	\frac{k_\delta}{k_0}=\Theta(1)exp(\frac{\Delta V_0-\Delta V_\delta}{\varepsilon})=\Theta(1)exp(\frac{\delta C(s_x)a^2+O(\delta^2)}{\varepsilon}),
\end{equation}
Where $\Theta(1)$ denotes a bounded proportional constant in the prefactor. Under the small-noise limit, this implies an exponential rate increase. We conducted numerical simulations of ground-state search for a 3-spin Ising model using the potential functions $V_1$ and $V_{\delta}$. By fitting the simulation data, we verified the above exponential acceleration effect. The comparison between the Hopfield network and cSB in terms of the average number of iterations required to reach the ground-state energy is shown below.


\begin{center}
\includegraphics[width=0.78\columnwidth]{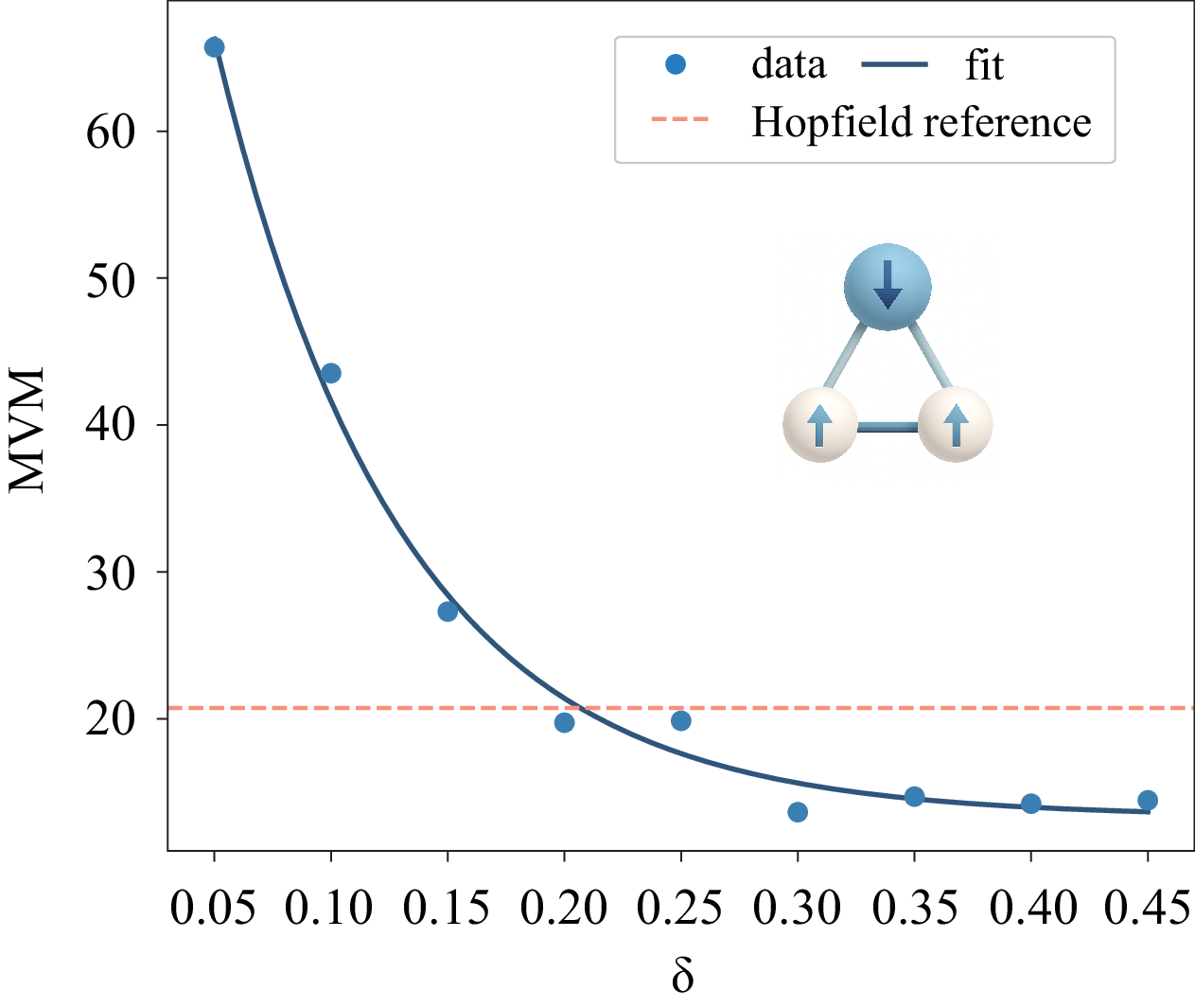}

\parbox{0.95\columnwidth}{\small \textbf{FIG. 5. Accelerated Ground State Search via Weak Coupling.} The relationship between the iterations of ground state solution for the 3-spin Ising model with the weak coupling strength $\delta$. The scatter points represent the iterations required for simulation at different $\delta$, the orange line represents the iterations in the Hopfield Network without weak coupling, and the blue curve represents the fitted result. All iterations are averaged over 100 runs. Trend line $log(mvm) = A\delta+B$, $A = -4.37$.
	}
\end{center}

\textit{cSB-EP update rules.} Then, we numerically solved the equations of the Hamiltonian system in the results section, with the cSB update rules satisfying
\begin{align}
	{y}_i(t+1) &= {y}_i(t)+\left\lbrace -x_i(t) + \rho\left[ \sum_{j=1}^N J_{i,j}x_j(t)\right] \right. \nonumber\\
	&\quad \left. -\gamma {y}_i(t)\right\rbrace \Delta_t,\\
	{x}_i(t+1) &= {x}_i(t) + y_i(t+1)\Delta_t.
\end{align}
Here, the activation function is $\rho(x) = \min\!\bigl(\max(x,0),1\bigr)$, $\Delta_t$ is the time step and we set $\Delta_t=1$.
After updating ${x}_i$, we will check whether $0<{x}_i< 1$. For the values exceeding this range, we will directly truncate ${x}_i$ and set ${y}_i=0$. This procedure effectively replaces the intricate computation involved in nonlinear barriers, while concurrently ensuring that the neuron's output remains constrained within the range of 0 to 1.

The cSB simulation equations give rise to the cSB-EP dynamics within the neural network's computational framework. For a fully connected neural network, the dynamics of the neurons $s_{i+1}$ in the $i+1$-th layer are governed by
\begin{align}
	sy_{i+1} &= (1-\gamma)sy_{i} - s_{i} + \rho\left[ W_is_{i}+h_i+W^T_{i+1}s_{i+2}\right],\\
	s_{i+1} &= s_{i} + sy_{i+1}.
\end{align}
Here, $W_i$ and $h_i$ represent the weight and bias of the $i$-th layer in the neural network, respectively. 

For the recognition output layer, the factor $\beta$ is crucial to align the output with the expected classification or prediction. It represents the target's influence on the output, effectively guiding the adjustment of weights and biases during training. Essentially, $\beta$ serves as a correction term that modulates the loss function $C$, thereby sharpening the relationship between the output and the target. 

The dynamics of the final layer neurons can be expressed as
\begin{align}
	sy_{end} &= (1-\gamma)sy_{end-1} - s_{end-1} + \rho\left[ W_{end}s_{end-1}+h_{end}\right] \nonumber\\
	&\quad +\beta(s_{end-1}-target),\\
	s_{end} &= s_{end} + sy_{end}.
\end{align}
By performing free-phase sampling, the steady state can be obtained. In this steady state, positive $\beta$ and negative $-\beta$ phase perturbations are applied separately, denoted as $s_t^{\beta}$ and $s_t^{-\beta}$. The gradient and bias of the network can be given by
\begin{equation}\begin{aligned}
		& -\frac{\partial C}{\partial W_{ij}}=\Delta W_{ij}=-\frac{1}{2\beta}\left(s_{i}^{\beta}s_{j}^{\beta}-s_{i}^{-\beta}s_{j}^{-\beta}\right) \\
		& -\frac{\partial C}{\partial h_{i}}=\Delta h_{i}=-\frac{1}{2\beta}\left(s_{i}^{\beta}-s_{i}^{-\beta}\right).
\end{aligned}\end{equation}

In adapting cSB-EP to CNNs, the matrix multiplication for weights is replaced by convolution. To align feature map dimensions across layers, up-pooling is applied before convolution. The remaining network components follow the same principles as in fully connected networks.
Our deep convolutional Hopfield network architecture aligns with that of ref. \cite{scellier2023energy}. It comprises an input layer, four convolutional layers, and a fully connected output layer with $10$ units corresponding to the number of categories in the task. Each convolutional layer uses a $3 \times 3$ kernel with padding of $1$, and the number of kernels is $3(1)$, $128$, $256$, $512$, and $512$, respectively, with $2\times2$ max pooling.

The energy function of the network could be expressed as:
\begin{equation}\begin{aligned}
		E(s,sy)&=\sum_{k=1}^{5}\frac{1}{2}\|s_{k}\|^{2}
		+\sum_{k=1}^{5}\frac{1}{2}\|sy_{k}\|^{2} \\
		&+\sum_{k=1}^{4}E_{k}^{\mathrm{conv}}(w_{k},s_{k-1},s_{k})
		+E_{5}^{\mathrm{fc}}(w_{5},s_{4},s_{5})\\
		&+\sum_{k=1}^{5}E_{k}^{\mathrm{bias}}(h_{k},s_{k}),
	\end{aligned}
\end{equation}
where $E_{k}^{\mathrm{conv}}=-s_{k}\bullet\mathcal{P}\left(w_k\star s_{k-1}\right)$, $E_{k}^{\mathrm{fc}}=-s_{k}^Tw_{k}s_{k}$, $E_{k}^{\mathrm{bias}}=-h_{k}^Ts_{k}$, $\mathcal{P}$ is the max pooling operation, and $\star$ denotes the convolution operation.

\begin{acknowledgments}
The authors gratefully acknowledge the support from the National Natural Science Foundation of China through Grants Nos.62401628, 62461160263, and 62371050.
\end{acknowledgments}

\bibliography{references}

\end{document}